\def\year{2017}\relax

\documentclass[letterpaper]{article}
\usepackage{aaai17}
\usepackage{helvet}
\usepackage{courier}
\usepackage{graphicx}
\usepackage[inline]{enumitem}
\usepackage{float}
\usepackage{subcaption}
\usepackage{color}

\usepackage{times}
\usepackage[hyphens]{url}

\title{Patch Reordering: a Novel Way to Achieve Rotation and Translation Invariance \\
       in Convolutional Neural Networks}

\author{Xu Shen $^{\dag}$, Xinmei Tian $^{\dag}$, Shaoyan Sun $^{\dag}$, Dacheng Tao $^{\ddag}$ \\
$^{\dag}$ CAS Key Laboratory of Technology in Geo-spatial Information Processing and Application System, \\ University of Science and Technology of China Hefei, Anhui, China 230027\\
$^{\ddag}$ Centre for Artificial Intelligence and the Faculty of Engineering and Information Technology, \\ University of Technology Sydney, 81 Broadway Street, Ultimo, NSW 2007, Australia}

\begin{document}

\maketitle

\begin{abstract}
Convolutional Neural Networks (CNNs) have demonstrated state-of-the-art performance on many visual recognition
tasks. However, the combination of convolution and pooling operations only shows invariance to small
local location changes in meaningful objects in input. Sometimes, such networks are trained using data augmentation
to encode this invariance into the parameters, which restricts the capacity of the model to learn
the content of these objects. A more efficient use of the parameter budget is
to encode rotation or translation invariance into the model architecture, which
relieves the model from the need to learn them. To enable the model to focus on
learning the content of objects other than their locations, we propose to
conduct patch ranking of the feature maps before feeding them into the next
layer. When patch ranking is combined with convolution and pooling operations, we obtain consistent
representations despite the location of meaningful objects in input. We show that the patch ranking module
improves the performance of the CNN on many benchmark tasks, including MNIST digit recognition, large-scale image
recognition, and image retrieval. The code is available at
\url{https://github.com//jasonustc/caffe-multigpu/tree/TICNN}.
\end{abstract}

\section{Introduction}
In recent years, convolutional neural networks (CNNs) have achieved state-of-the-art performance on
many computer vision tasks, including image recognition \cite{GoogleNet,VGG,ResNet},
semantic segmentation \cite{FCNN}, image captioning \cite{Feifei-Image-Cap,Jeff-Image-Cap,MS-Image-Cap},
action recognition \cite{action-rcnn,action-two-stream}, and video captioning \cite{yaoli-video-caption,msra-video-caption}.
The success of CNNs comes from their ability to learn the two-dimensional structures of images for which
objects and patterns may appear at different locations. To detect and learn patterns despite their locations,
the weights of local filters are shared when applied to different positions in the image.

\begin{figure*}
    \centering
    \begin{subfigure}[t]{0.45\linewidth}
        \includegraphics[width=\linewidth]{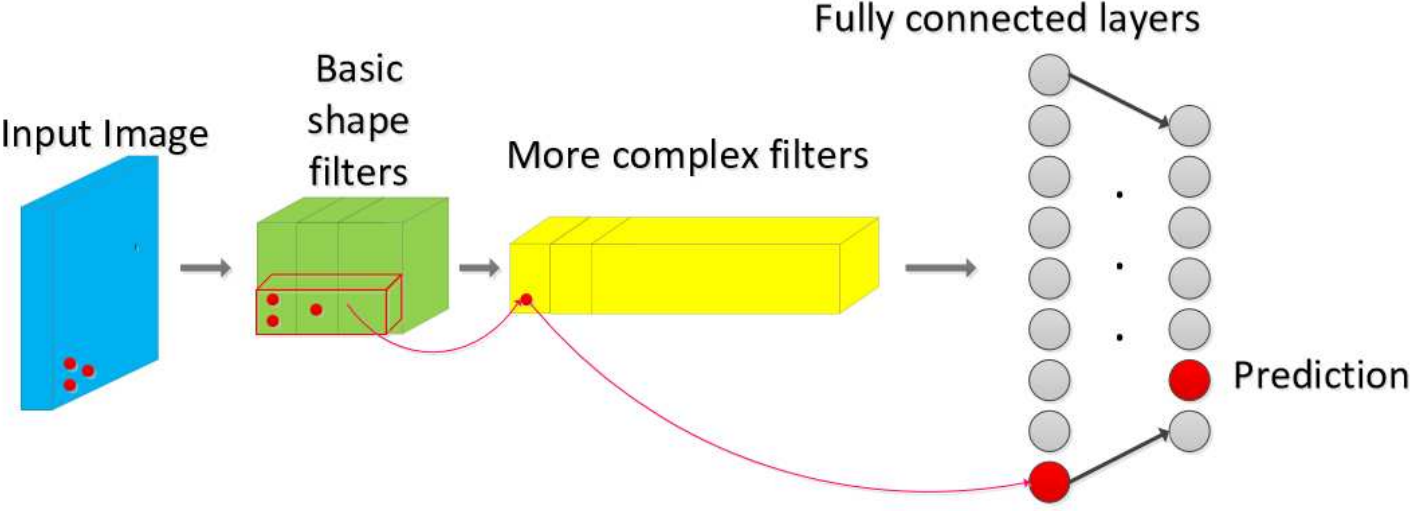}
    \end{subfigure}
    \hspace{10mm}
    \begin{subfigure}[t]{0.45\linewidth}
        \includegraphics[width=\linewidth]{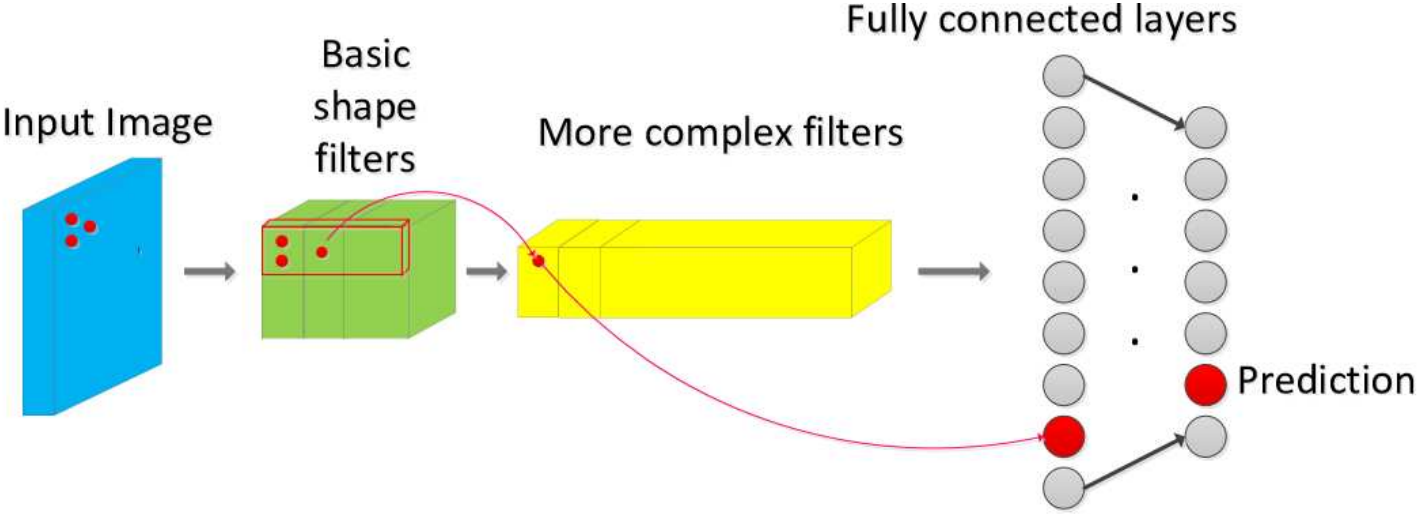}
    \end{subfigure}
    \caption{Large global location changes of patterns in input (e.g., global
        rotation or translation of objects) will propagate
        to the feature maps of the final convolutional layer. As a result, the
        fully connected layers have to encode the invariance of location changes into
        its parameters, which restricts the capacity of the model to learn the
        content of these objects.}
    \label{fig:propagation}
\end{figure*}

Since distortions or shifts of the input can cause the positions of salient features to vary, weight sharing is very important for
CNNs to detect invariant elementary features regardless of location changes of these features \cite{Lecun-cnn}. In addition,
pooling also reduces the sensitivity of the output to small local shifts and distortions by reducing the resolutions of
input feature maps. However, another important property of weight sharing or pooling is that the location of detected features
in the output feature maps is identical to that of the corresponding local patch in the input feature maps. As a result,
the location change of the input visual patterns in lower layers will propagate to higher convolutional layers. Due to the typically small local spatial support for pooling (e.g., $2\times2$) and convolution (e.g., $9\times9$ kernel size), large global location changes of patterns in input (e.g., global rotation or translation of objects) will even propagate to the feature maps of the final convolutional layer (as shown in Fig. \ref{fig:propagation}). Consequently, the following fully connected layers have to learn the location invariance to produce consistent predictions or representations, which restricts the use of the parameter budget for achieving more powerful outputs.

In this paper, we introduce a \textit{Patch Reordering (PR)} module that can be embedded into a standard CNN architecture to improve the rotation and translation invariance capabilities. Output feature maps of the convolutional layer are first divided into multiple tiers of non-overlapped local patches at different spatial pyramid levels. We reorder these local patches at each level based on their energy (e.g., L1 or L2 norm of activations of the patch). To retain the spatial consistency of local patterns, we only reorder the patches of a given level locally (i.e., within each single patch of its upper level). In convolutional layers, a location change of the patterns in input feature maps will result in a corresponding location change of the output feature maps, while the local patterns (activations) in the output are equivalent. As a result, ranking these local patterns in a specific order leads to a consistent representation despite the locations of local patterns in input, that is, rotation or translation invariance. The proposed architecture can be inserted into any convolutional layers and allows for end-to-end training of the models for which they are applied. In addition, we do not need any extra training supervision or modification to the training process or any preprocessing of input images.

\section{Related Work}

The equivalence and invariance of CNN representations to input image transformations were investigated in  \cite{cvpr-invest-inva,iclr-invest-inva,nips-invest-inva}. Specifically, Cohen and Welling \cite{iclr-invest-inva}
showed that a linear transform of a good visual representation was equivalent to a combination of the elementary
irreducible representations using the theory of group representations. Lenc and Vedaldi \cite{cvpr-invest-inva}
estimated the linear relationships between representations of the original and transformed images. Gens and Domingos
\cite{nips-invest-inva} proposed a generalization of CNNs that formed feature maps over arbitrary symmetry groups
based on the theory of symmetry groups in \cite{cvpr-invest-inva}, resulting in feature maps that were more invariant
to symmetry groups. Bruna and Mallat \cite{pami-invariant} proposed a wavelet scattering network to compute a translation invariant image representation. Local linear transformations were adopted in the feature learning algorithms in \cite{icml-local-trans} for the purpose of transformation-invariant feature learning.

Numerous recent works have focused on introducing spatial invariance in deep learning architectures explicitly.
For unsupervised feature learning, Sohn and Lee \cite{icml-local-trans} presented a transform-invariant restricted Boltzmann
machine that compactly represented data by its weights and their transformations, which achieved invariance of the feature
representation via probabilistic max pooling. Each hidden unit was augmented with a latent transformation assignment variable
that described the selection of the transformed view of the weights associated with the unit in \cite{inva-rbm}. In both works,
the transformed filters were only applied at the center of the largest receptive field size. In tied convolutional
neural networks \cite{tiled-cnn}, invariance was learned explicitly by square-root pooling hidden units computed by
partially un-tied weights. Here, additional learned parameters were needed when un-tying weights.

The latest two works on incorporating spatial invariance in CNNs are described in \cite{sci-cnn,spatial-trans-cnn}.
In \cite{sci-cnn}, feature maps in CNNs were scaled or rotated to multiple levels, and the same kernel was convolved across the input at each scale. Then, the responses of the convolution at each scale were normalized and pooled at each spatial location to obtain a locally scale-invariant representation. In this model, only limited scales were considered, and extra modules were needed in the feature extraction process. To address different transformation types in input images, Jaderberg \emph{et al.} \cite{spatial-trans-cnn} proposed inserting a spatial transformer module between CNN layers, which explicitly transformed an input image into a proper appearance and fed the transformed input into the CNN model.


In conclusion, all aforementioned related works improve the transform invariance of deep learning models by adding
\textit{extra feature extraction modules}, \textit{more learnable parameters}, or \textit{extra transformations on input images},
which makes the trained CNN model problem-dependent and not generalizable to other datasets. In contrast, in this paper, we propose a very simple reordering on feature maps during the training of CNN models. No extra feature extraction modules or more learnable parameters are needed. Therefore, it is very easy to apply the trained model to other vision tasks.

\section{Patch Reordering in Convolutional Neural Networks}
Weight sharing in CNNs allows feature detectors to detect features regardless of their spatial locations in the image; however, the corresponding location of output patterns varies when subject to location changes of the local patterns in the input. Learning invariant representations causes parameter redundancy problems in current CNN models. In this section, we will reveal this phenomenon and propose the formulation of our \textit{Patch Reordering} module.

\subsection{Parameter Redundancy in Convolutional Neural Networks}
\begin{figure*}
\centering
    \begin{subfigure}[t]{0.45\linewidth}
        \includegraphics[width=\textwidth]{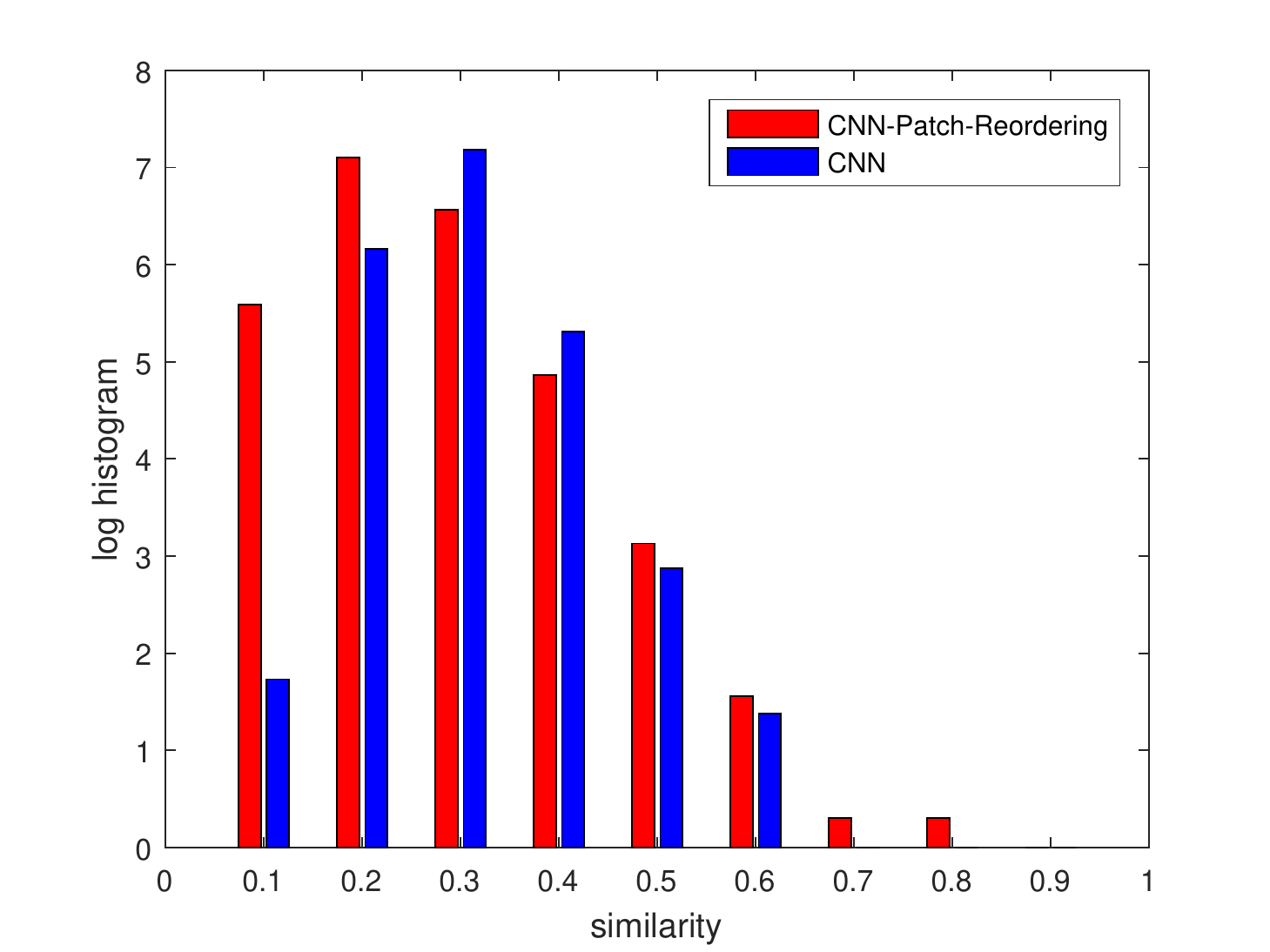}
        \caption{Similarity of weights in fc6.}
    \end{subfigure}
    \hspace{10mm}
    \begin{subfigure}[t]{0.45\linewidth}
        \includegraphics[width=\textwidth]{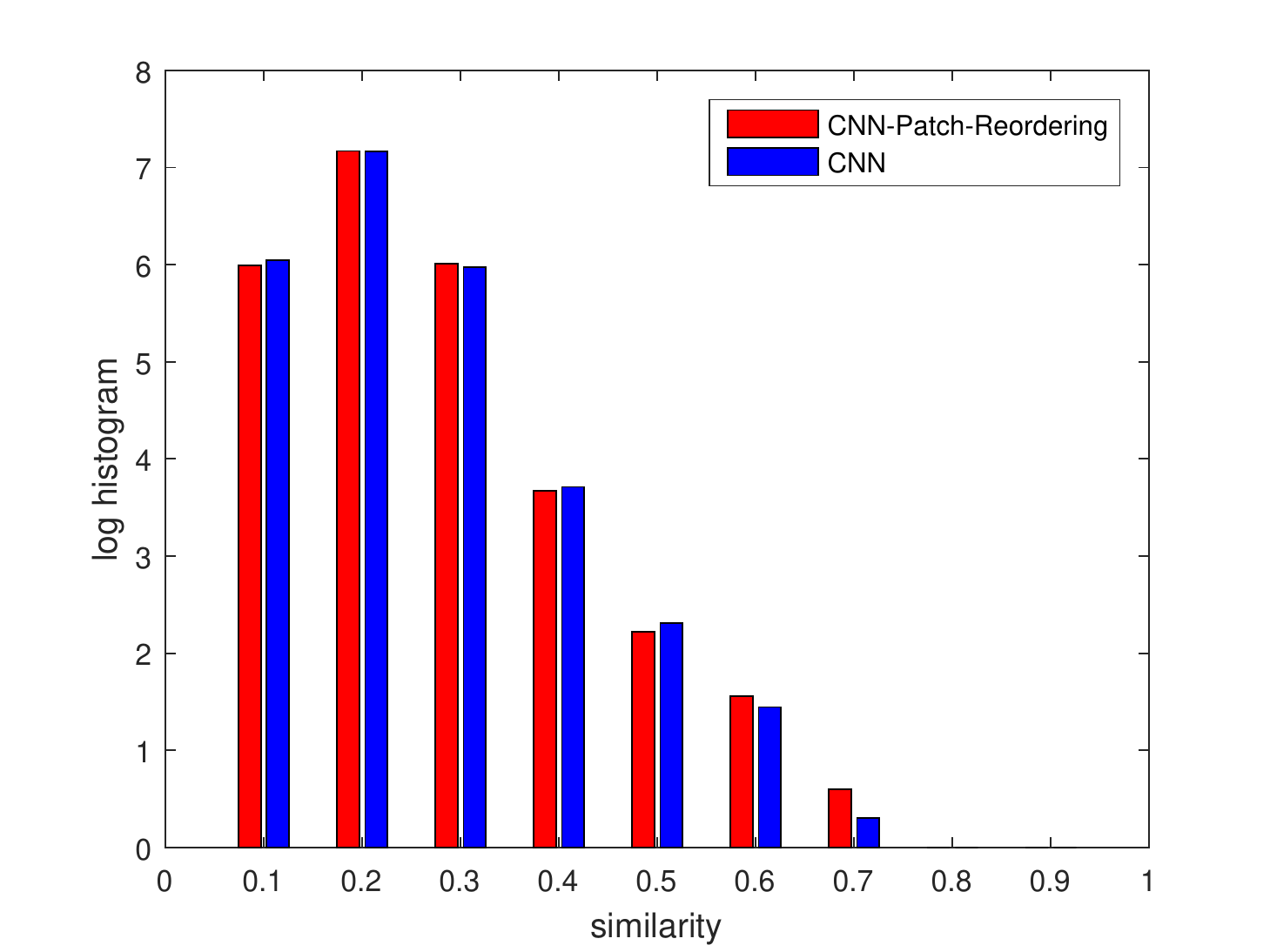}
        \caption{Similarity of weights in fc7.}
    \end{subfigure}
    \vspace{-3mm}
\caption{Log histogram of similarity between weights in fc6 (a) and fc7 (b) layers in a CNN with (CNN-Patch-Reordering) and without the (CNN) patch reordering module. In conventional CNNs, the correlation of parameters in fc6 is much higher than that in fc7, while that for PR-CNN is quite consistent. This phenomenon implies that the location change of input visual patterns leads to a higher parameter redundancy in the subsequent layers.}
\label{fig:loc_change}
\end{figure*}

Let $X^{l-1}=\{\vec{x}_{ij}, i=1,\cdots,h, j=1,\cdots,w\}$ denote the output feature maps of a convolutional layer with $m$ elements ($m= h\times w \times c$ for $c$ feature maps with height $h$ and width $w$). Each $\vec{x}_{ij}$ is a $c$-dimensional input feature vector corresponding to location ($i$,$j$). If it is followed by a fully connected layer, $X^l$ can be computed by
\begin{equation}
X^l = f(W X^{l-1}) = f(\sum_{i=1}^{h} \sum_{j=1}^{w} \vec{w}_{ij}^T \vec{x}_{ij}),
\end{equation}
where $f$ is a non-linear activation function and $\vec{w}_{ij}$ are the weights for location ($i$,$j$). If there is some location change (such as a rotation or translation) of the input features, the resulting new input becomes $X'^{l-1}=\{\vec{x'}_{i'j'}, i'=1,\cdots,h, j'=1,\cdots,w\}$. Since there are no value changes (except cropping or padding), for $\vec{x}_{ij}$ in any position ($i$,$j$), we can always find its correspondence $\vec{x'}_{i'j'}$ in the transformed input, i.e. $\vec{x}_{ij}=\vec{x'}_{i'j'}$. If the network learns to be invariant under this type of location change, the output (or representation) should remain the same. Specifically,

\begin{equation}
X'^l \approx X^l \Rightarrow f(\sum_{i'=1}^{h} \sum_{j'=1}^{w} \vec{w}_{i'j'}^T \vec{x'}_{i'j'}) \approx f(\sum_{i=1}^{h} \sum_{j=1}^{w} \vec{w}_{ij}^T \vec{x}_{ij}) ~.
\end{equation}

Then, in the monotonous section of $f$, we have
\begin{equation}
\sum_{i'=1}^{h} \sum_{j'=1}^{w} \vec{w}_{i'j'}^T \vec{x'}_{i'j'} \approx \sum_{i=1}^{h} \sum_{j=1}^{w} \vec{w}_{ij}^T \vec{x}_{ij} ~.
\end{equation}

Since $\vec{x}_{ij}=\vec{x'}_{i'j'}$, the aforementioned equation can be simplified as:
 \begin{equation}
 \label{conclusion}
 \sum_{i'=1}^{h} \sum_{j'=1}^{w} (\vec{w}_{i'j'}-\vec{w}_{ij})^T \vec{x}_{ij} \approx \vec{0}
\end{equation}

Because $\vec{x}_{ij}$ varies as the input image changes, we have $\vec{w}_{i'j'}-\vec{w}_{ij}\approx \vec{0}$. That is to say, encoding rotation or translation invariance into CNNs leads to highly correlated parameters in higher layers. Therefore, the capacity of the model decreases.
To validate this redundancy in CNN models, we compare the log histogram of cosine similarities between weights in fc6 and fc7 in AlexNet \cite{AlexNet}. Fig. \ref{fig:loc_change} shows that parameter redundancy of the model is significantly reduced because of a more consistent feature map after patch reordering.

\subsection{Patch Reordering}

\begin{figure*}
\centering
    \begin{subfigure}[t]{0.25\linewidth}
        \includegraphics[width=\textwidth]{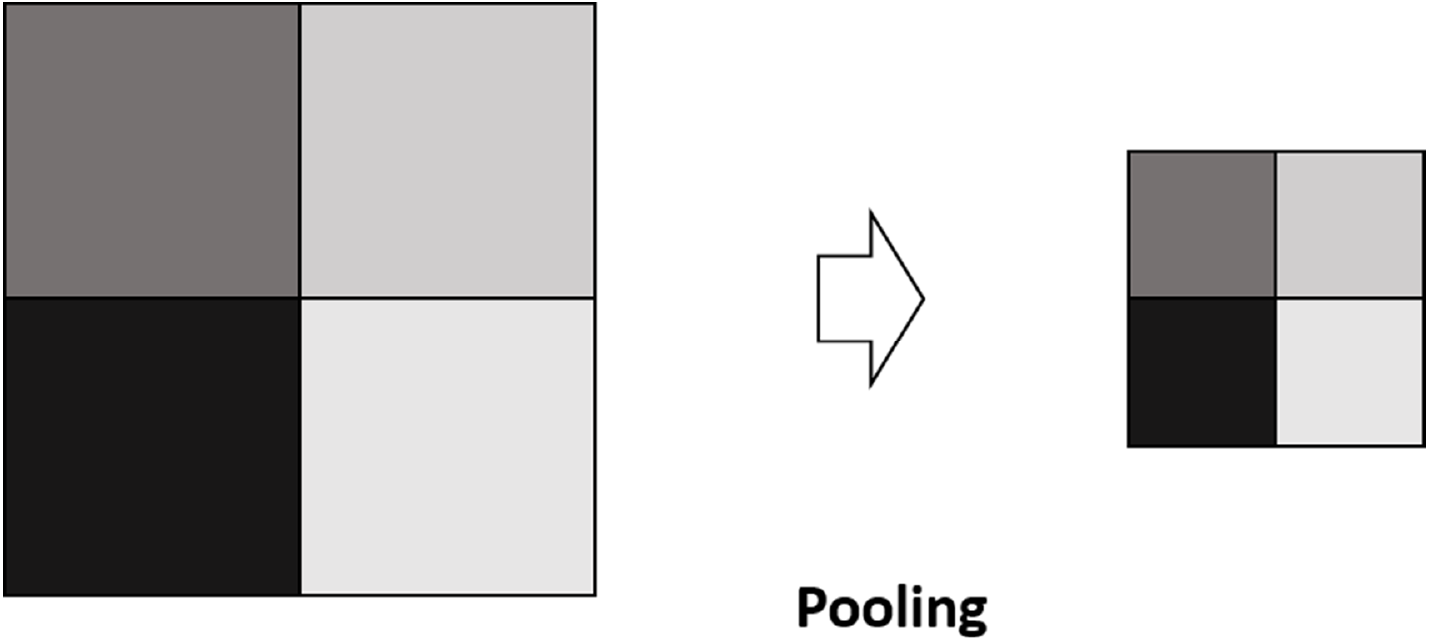}
        \caption{Pooling layer.}
    \end{subfigure}
    \hspace{15mm}
    \begin{subfigure}[t]{0.65\linewidth}
        \includegraphics[width=\textwidth]{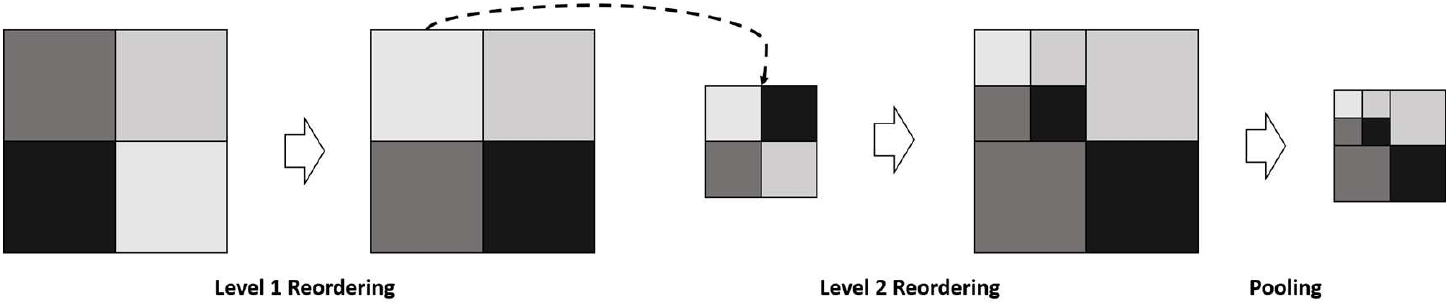}
        \caption{Pooling layer with patch reordering.}
    \end{subfigure}

\caption{Side-by-side comparison of the structure of (a) a conventional convolutional layer and (b) the proposed convolutional layer with patch reordering module. Feature maps are divided into $(n\times n)^l$ non-overlapped patches at level-$l$. Here, we take $n=2$ and $l=2$ as an example. The four patches of level $1$ are first reordered so that patches with higher energy precedes other patches. Then, we repeat this process within each single patch in the previous level (here, we only show the reordering of the first patch in level 2).  One visual example of the output feature maps between original and rotated/translated input images with/without patch reordering is illustrate in Supplemental Material.}
\label{fig:framework}
\end{figure*}

If one object is located at different positions in two images, the same visual features of the object will locate at different positions in their corresponding convolution feature maps. The feature maps generated by deep convolutional layers are analogous to the feature maps in traditional methods \cite{encoding,quantization}. In those methods, image patches or SIFT vectors are densely extracted and then encoded. These encoded features compose the feature maps and are pooled into a histogram of bins. Reordering of the pooled histogram achieves translation and rotation invariance. Likewise, since the deep convolutional feature maps are the encoded representations of images, reordering can be applied in a similar way.

Since convolutional kernels function as feature detectors, each activation in the output feature maps corresponds to a match of a specific visual pattern. Therefore, when the feature detectors slide through the whole input feature maps, the locations with matched patterns generate very high responses and \emph{vice versa}. Consequently, the ``energy'' distribution (L1 norm or L2 norm) of the local patches in the output feature maps presents some heterogeneity. Furthermore, patches with different energies correspond to different parts of the input object. Naturally, if we rank the patches by their energies in a descending or ascending order, regardless of how we change the location of visual patterns by rotation or translation in the input, the output order will be quite consistent. Finally, and rotation- and translation-invariant representation is generated.

\subsubsection{Forward Propagation}

The details of the patch reordering module are illustrated in Fig. \ref{fig:framework}. The feature maps are divided into $(n\times n)^l$ non-overlapped patches at level-$l$. Here, $n$ is a predefined parameter (e.g., $2$ or $3$). Then, we rank the patches by energy (L1 or L2 norm) within each patch of level $l-1$:
\begin{equation}
En^{l-1}_{i\in(w^{l-1}_1, w^{l-1}_2); j\in(h^{l-1}_1, h^{l-1}_2)} = \sum_{i=w^{l-1}_1}^{w^{l-1}_2}\sum_{j=h^{l-1}_1}^{h^{l-1}_2} |X_{i,j}| ~or ~X_{i,j}^2~.
\end{equation}
The patches are located from the upper left to the lower right in descending order of energy. The offset of each pixel in the patch $(o^l_i,o^l_j)$ can be obtained from the gap between the target patch location and the source patch location. Finally, the output feature map can be computed by
\begin{equation}
Z_{i,j} = X_{i + \sum_{k=1}^l o^k_i, ~ j + \sum_{k=1}^l o^k_j}~.
\end{equation}

\subsubsection{Backward Propagation}
During the back-propagation process, we simply pass the error from the output pixel to its corresponding input pixel:
\begin{equation}
E^X_{i,j} = E^Z_{i - \sum_{k=1}^l o^k_i, ~ j - \sum_{k=1}^l o^k_j}~.
\end{equation}

\section{Experiments} \label{sec:exp}
In this section, we evaluate our proposed CNN with patch reordering module on several supervised learning tasks, and compare our model with state-of-the-art methods, including traditional CNNs, SI-CNN \cite{sci-cnn}, and ST-CNN \cite{spatial-trans-cnn}. First, we conduct experiments on the distorted versions of the MNIST handwriting dataset as in \cite{spatial-trans-cnn,sci-cnn}. The experimental results show that patch reordering is capable of achieving comparable or better classification performance. Second, to test the effectiveness of patch reordering on CNNs for large-scale real-world image recognition tasks, we compare our model with AlexNet \cite{AlexNet} on ImageNet-2012 dataset. The results demonstrate that patch reordering improves the learning capacity of the model and encodes translation and rotation invariance into the architecture even when trained on raw images only. Finally, to evaluate the generalization ability of the proposed model on other vision tasks with real-world transformations of images, we apply our model to solve the image retrieval task on UK-Bench \cite{uk-bench} dataset. The improvement in the retrieval performance reveals that the proposed model has a good generalization ability and is better at solving real-world transformation variations.

We implement our method using the open-source Caffe framework \cite{caffe}. For patch energy, we have tested both L1 and L2 norm and found that they did not show much difference. Our code and model will be available online.
For SI-CNN and ST-CNN, we directly report their results from the original papers on MNIST. For ImageNet-2012, since these two methods did not report their results on this dataset, we forked from the github for re-implementation.

\subsection{MNIST}\label{sec:mnist}
In this section, we use the MNIST handwriting dataset to evaluate all deep models. In particular, different neural networks
are trained to classify MNIST data that have been transformed via rotation (R) and translation (T). The rotated dataset was generated from rotating digits
with a random angle sampled from a uniform distribution $U[-90^{\circ}, 90^{\circ}]$. The translated dataset was generated by randomly locating the $28\times28$ digit in a $42\times42$ canvas.

\begin{table}[t!]
\begin{center}
\begin{tabular}{|l|c|c|}
\hline
Method & R & T \\
\hline
FCN & 2.1 & 2.9 \\
\hline
CNN & 1.2 & 1.3 \\
\hline
SI-CNN  & 0.9 & -  \\
\hline
ST-CNN & \textbf{0.8} &  0.8 \\
\hline
PR-CNN(ours) & \textbf{0.8} & \textbf{0.7} \\
\hline
\end{tabular}
\end{center}
\vspace{-4mm}
\caption{Classification error on the transformed MNIST dataset. The different distorted MNIST datasets
are R (rotation) and T (translation). All models have the same number of parameters and use the same basic architectures.}
\vspace{-4mm}
\label{perf:mnist}
\end{table}

Following \cite{spatial-trans-cnn}, all networks use ReLU activation function and softmax classifiers. All CNN networks have a $9\times9$ convolutional layer (stride $1$, no padding), a $2\times2$ max-pooling layer with stride $2$, a subsequent $7\times7$ convolutional layer (stride $1$, no padding), and another $2\times2$ max-pooling layer with
stride $2$ before the final classification layer. All CNN networks have $64$ filters per layer. For SI-CNN, convolutional layers are replaced by rotation-invariant layers using six angles from $-90^\circ$ to $90^\circ$. For ST-CNN, the spatial transformer module is placed at the beginning of the network. In our patch reordering CNN, the patch reorder module is applied to the second convolutional layer. The feature maps are divided into $4^l$ blocks at level $l$. Here, we set $l=1$.
All networks are trained with SGD for $150000$ iterations, with a $256$ batch size, $0.01$ base learning rate, and no weight decay or dropout. The learning rate was reduced by a factor of $0.1$ every $50000$ iterations. Weights were initialized randomly,
and all networks shared the same random seed.

The experimental results are summarized in Table \ref{perf:mnist}. It shows that our model achieves better performance under translation and comparable performance under rotation. Because our model does not need any extra learnable parameters, feature extraction modules, or transformations on training images, the comparable performance still reflects the superiority of the patch reordering CNN. For ST-CNN, the best results reported in \cite{spatial-trans-cnn} is obtained by training with a more narrow class of transformations selected manually (affine transformations). In our method, we did not optimize with respected to transformation classes. Therefore the comparison is unfair for our PR-CNN. We should compare with the most general ST-CNN defined for a class of projection transformations: 0.8(R) and 0.8(T).

\subsection{ImageNet-2012}\label{sec:imagenet}
\begin{table}[t!]
\begin{center}
\begin{tabular}{|l|c|c|c|}
\hline
Method & Ori & R & T \\
\hline
CNN & 57.1/80.2 & 36.6/57.7 & 46.5/70.8 \\
\hline
CNN-Data-Aug & 56.6/79.8 & 36.5/58.3 & 50.0/73.9 \\
\hline
SI-CNN  & 57.2/80.2 & 36.8/58.9 & - \\
\hline
ST-CNN  & 59.1/81.7 & 37.3/59.3 & 51.4/75.3 \\
\hline
PR-CNN(ours) &  \textbf{60.4/82.4} & \textbf{40.7/63.3} & \textbf{54.9/78.0} \\
\hline
\end{tabular}
\end{center}
\vspace{-3mm}
\caption{Classification accuracy on the ImageNet-2012 validation dataset. The evaluation datasets are Ori (original), R (rotation), and T (translation). }
\label{perf:imagenet}
\end{table}
The ImageNet-2012 dataset consists of images from $1000$ classes and is split into three subsets: training ($1.3$M),
validation ($50$K), and testing ($100$K images with held-out class labels). The classification
performance is evaluated using the top-1 and top-5 accuracy. The former is a multi-class
classification accuracy. The latter is the main evaluation criterion used in ILSVRC and is defined
as the proportion of images whose ground-truth category is not in the top-5 predicted categories.
We use this dataset to test the performance of our model on a large-scale image recognition task.

CNN models are trained on raw images and tested on both raw and transformed images. For all transform types, specific
transformations are applied to the original images. Then, the transformed images are rescaled to have a smallest image side of  $256$ pixels. Finally, the center $224\times224$ crop is used for test. The rotated (R) dataset
is generated by randomly rotating original images from $-45^{\circ}$ to $45^{\circ}$ with a uniform distribution. The translated dataset (T) is generated by randomly shifting an image by a proportion of $U[-0.2, 0.2]$.

All the models follow the architecture of AlexNet. For SI-CNN, the first, second and fifth convolutional layers are replaced by rotation-invariant layers using six angles from $-90^\circ$ to $90^\circ$. For ST-CNN, the input is fed into a spatial transformer network before the AlexNet. The spatial transformer network uses bilinear sampling with an affine transformation function. As in \cite{spatial-trans-cnn}, the size of the spatial transformer network is about half the size of  AlexNet. For our PR-CNN, the feature maps are divided into $4^l$ blocks at level $l=2$, and the
patch reorder module is applied to the fifth convolutional layer.

To train SI-CNN and PR-CNN, we use a base learning rate of $0.01$ and decay it by $0.1$ every
$200,000$ iterations. Both networks are trained for $700,000$ iterations. We use a momentum of $0.9$, a weight decay of $0.0005$, and a weight clip of $35$. The convolutional
kernel weights and bias are initialized by $\mathcal{N}(0, 0.01^2)$ and $0.1$, respectively. The weights and bias of
fully connected layers are initialized by $\mathcal{N}(0, 0.005^2)$ and $0.1$. The bias learning rate is set to be \
$2\times$ the learning rate for the weights. For ST-CNN, since it does not converge under the aforementioned setting, we fine-tune the network with the classification network
initialized by the pre-trained AlexNet.
The spatial transformer module consists of $2$ convolutional layers, $2$ pooling layers, and $3$ fully connected layers. The first convolutional layer filters the input with $48$ kernels of size $11\times11\times3$ with a stride of $4$ pixels, then is connected by a $3\times3$ pooling layer with stride $2$. The second convolutional layer has $48$ kernels of size $5\times5\times48$ with a stride of $2$ pixels, followed by a $3\times3$ max pooling layer with stride $2$. The output of the pooling layer is fed into two fully connected layers with $48$ neurons. Finally, the third fully connected layer maps the output into $6$ affine parameters. Then the 6-dimensional output is fed into the spatial transformer layer to get the transformed input image.
During the fine-tuning, the learning rate of the spatial transformer is set to be $0.01\times$ that of the classification network. We use a base learning rate
of $0.001$ and decay it by $0.1$ every $50,000$ iterations, the training process converges after approximately $200,000$ iterations.

The results are presented in Table \ref{perf:imagenet}. It shows that data augmentation, feature map augmentation, transform pre-processing and patch reordering are all effective ways to improve the rotation or translation invariance of CNNs. Our PR-CNN not only achieves more consistent representation faced with location changes in input but also relieves the models from encoding invariance. It
improves the classification accuracy of the model even for the original test images.

\subsection{UK-Bench}\label{sec:uk-bench}
\begin{table}[t!]
\begin{center}
\begin{tabular}{|l|c|c|}
\hline
Method & FC6 & FC7 \\
\hline
CNN & 3.381 & 3.438 \\
\hline
CNN-Data-Aug & 3.340 & 3.441 \\
\hline
SI-CNN  & 3.431 & 3.452 \\
\hline
ST-CNN  & 3.430 & 3.446 \\
\hline
PR-CNN(ours) &  \textbf{3.574} & \textbf{3.539} \\
\hline
\end{tabular}
\end{center}
\vspace{-3mm}
\caption{Performance of CNN models on the UK-Bench retrieval dataset. Here, we use the $4096$-dimensional feature of fc6 and fc7 for evaluation.}
\label{perf:uk-bench}
\end{table}
We also evaluate our PR-CNN model on the popular image retrieval benchmark dataset UK-Bench \cite{uk-bench}. This dataset includes $2550$ groups of images, each containing $4$ relevant samples concerning a certain object or scene from different viewpoints. Each of the in total $10200$ images is used as one query to perform image retrieval, targeting at finding each image's $3$ counterparts. We choose UK-Bench since the viewpoint variation in the dataset is very common. Although many of the variation types are beyond the three types of geometry transformations that we attempt to address with, we demonstrate the effectiveness of PR-CNN for solving many severe rotation, translation and scale variance cases in image retrieval task.

We directly apply the models trained on ImageNet-2012 for evaluation. The outputs of the fc6 and fc7 layers are used as the feature for each image. Then, we compute the root value of each dimension and perform L2 normalization.
To perform image retrieval on UK-Bench, the Euclidean distances of the query image with respect to all $10200$ database images are computed and sorted. Images with the smallest distances are returned as top ranked images. NS-Score (average top four accuracy) is used to evaluate the performance, and a score of $4.0$ indicates that all the relevant images are successfully retrieved in the top-four results.

As shown in Table \ref{perf:uk-bench}, data augmentation, feature map augmentation or spatial transformer network does not present considerable capacity of transform invariance when applied to an unrelated new task. Maybe these models need to be well fine-tuned when transferred to a new dataset and the Spatial Transformer block is content and task dependent. Patch reordering is better for transferring by encoding invariance only into architecture, which is irrelevant to the content of input. It demonstrates that our PR-CNN model can be seamlessly transferred to other image recognition based applications (e.g. image retrieval) without any re-training/fine-tuning. Meanwhile, for other models, fc7 presents better invariance than fc6. However, for our PR-CNN, fc6 is better. We can find some clues from Fig. \ref{fig:loc_change}, that is, fc6 presents less parameter redundancy than fc7 in PR-CNN.

\subsection{Measuring Invariance}

\begin{figure}
\centering
\includegraphics[width=0.45\textwidth]{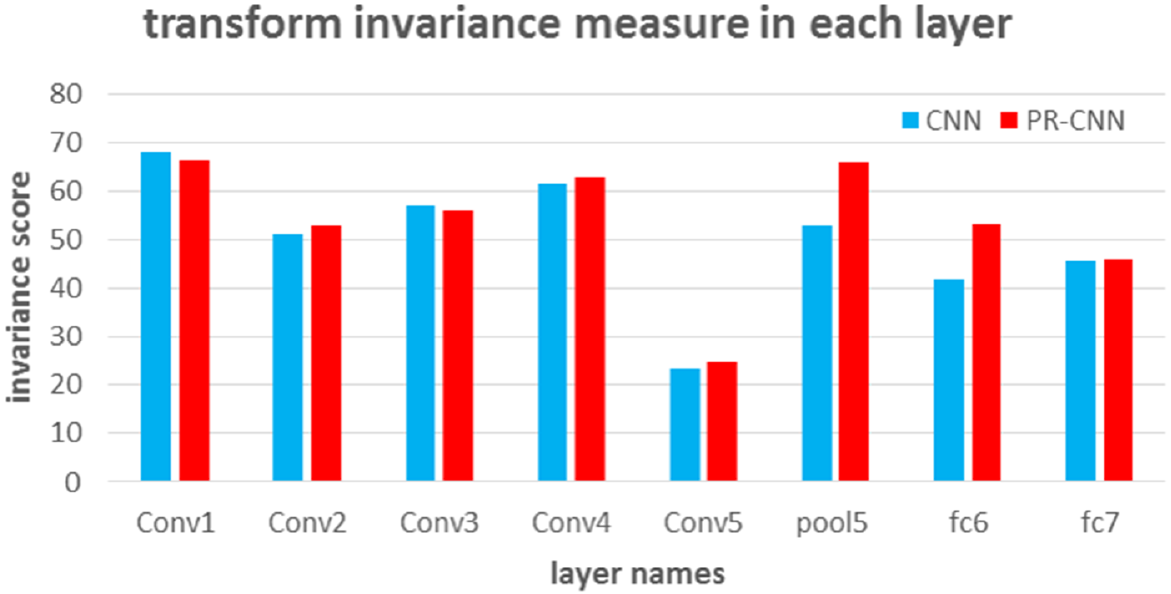}
    \caption{Transform invariance measure (the larger, the better). By applying
    patch reordering on feature maps during training, the invariance of the
    following layers is significantly improved.}
    \label{fig:invariance-measure}
\end{figure}

We evaluate the transform invariance achieved by our model using the invariance measure proposed in \cite{invariance-measure}.
In this approach, a neuron is considered to be firing when its response is above a certain threshold $t_i$. Each $t_i$ is chosen
to satisfy the condition that $G(i) = \sum |h_i(x) > t_i| / N$ is greater than $0.01$, where $N$ is the number of inputs.
Then, the \textit{local firing rate} $L(i)$ is computed as the proportion of transformed inputs to which a neuron fires. To ensure that
a neuron is selective and with a high local firing rate (invariance to the set of the transformed inputs), the invariance score of a
neuron is computed based on the ratio of its invariance to selectivity, i.e., $L(i) / G(i)$. We report the average of the top $20\%$
highest scoring neurons ($p=0.2$), as in \cite{sci-cnn}. Please refer to \cite{invariance-measure} for more details.

Here, we build the transformed dataset by applying rotation ($[-45^{\circ}, 45^{\circ}]$ with a step size of $9^{\circ}$) and translation ($[-0.2, 0.2]$ with a step size of $0.04$) on the $50,000$ validation images
of ImageNet-2012. Fig. \ref{fig:invariance-measure} shows the invariance score of CNN and PR-CNN measured at the end of each layer. We can see that by applying patch reordering to feature maps during training, the invariance of the subsequent layers is significantly improved.

\begin{figure}[t]
\centering
    \begin{subfigure}[t]{0.9\linewidth}
        \includegraphics[width=\textwidth]{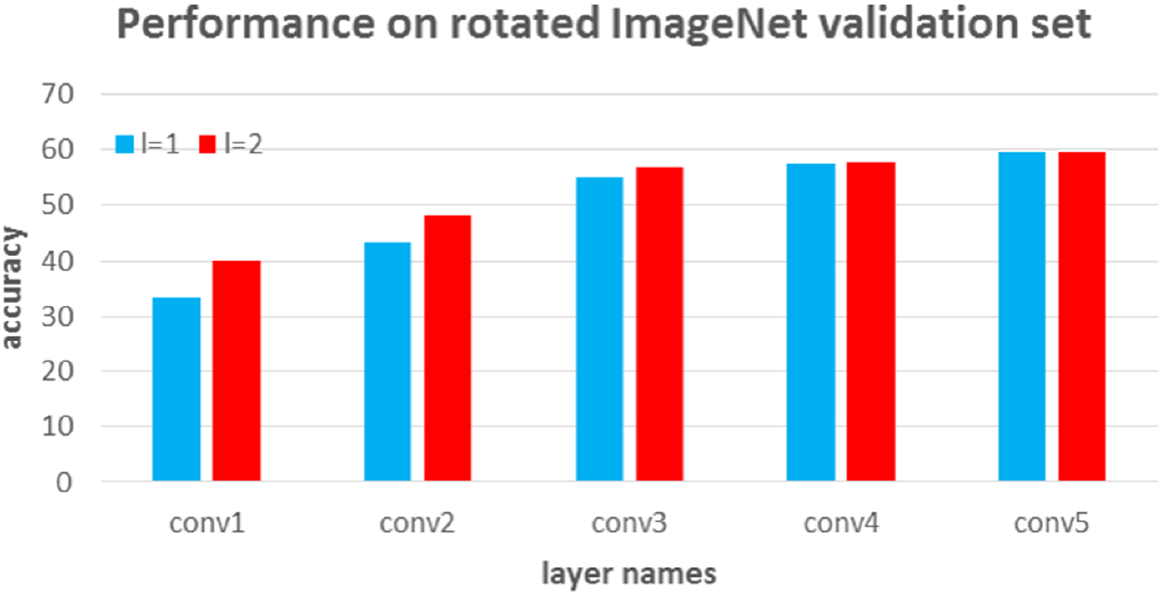}
        \caption{Performance of PRCNN for rotated images.}
    \end{subfigure}
    \begin{subfigure}[t]{0.9\linewidth}
        \includegraphics[width=\textwidth]{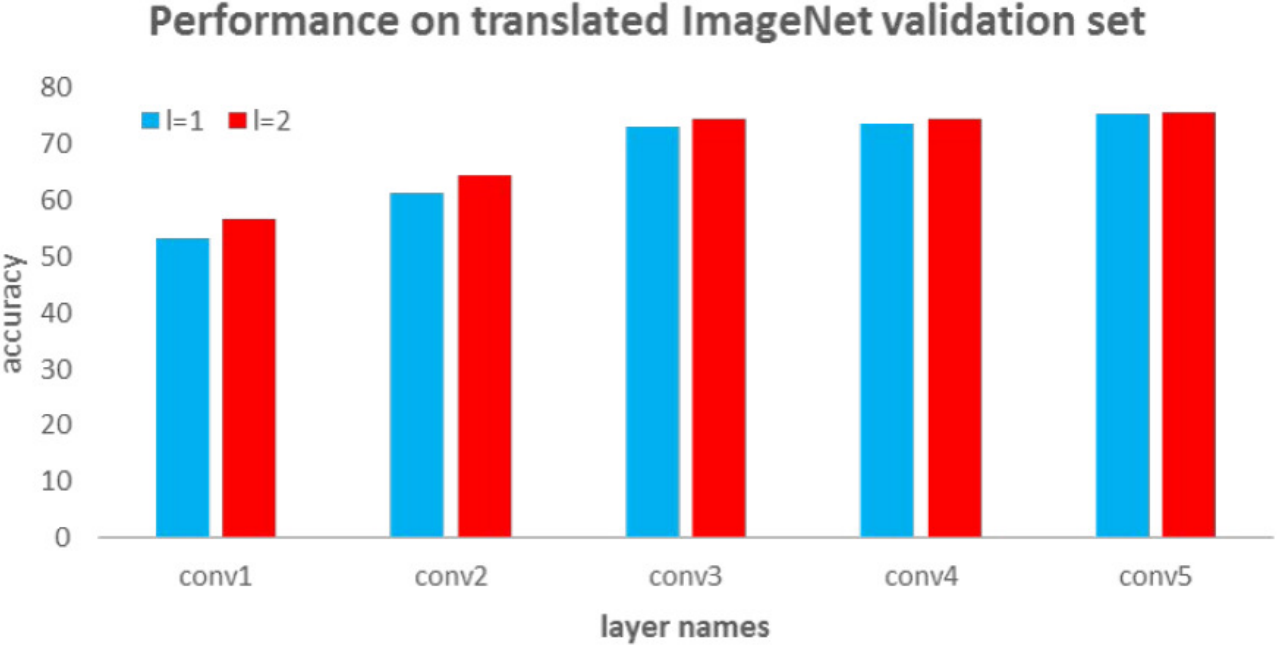}
        \caption{Performance of PRCNN for translated images.}
    \end{subfigure}
    \caption{Performance of PRCNN when the PR module is applied to different
    layers with different levels. The performance drops significantly when we
    perform patch reordering in low layers. This is because patch reordering breaks
    local spatial correlations among patches, which is vital for recognizing
    meaningful visual patterns in lower layers.}
\label{fig:effect}
\end{figure}

\subsection{Effect of Patch Reordering on Image Representations}
\begin{figure*}
    \centering
    \begin{subfigure}[t]{\linewidth}
        \centering
        \includegraphics[width=0.9\linewidth]{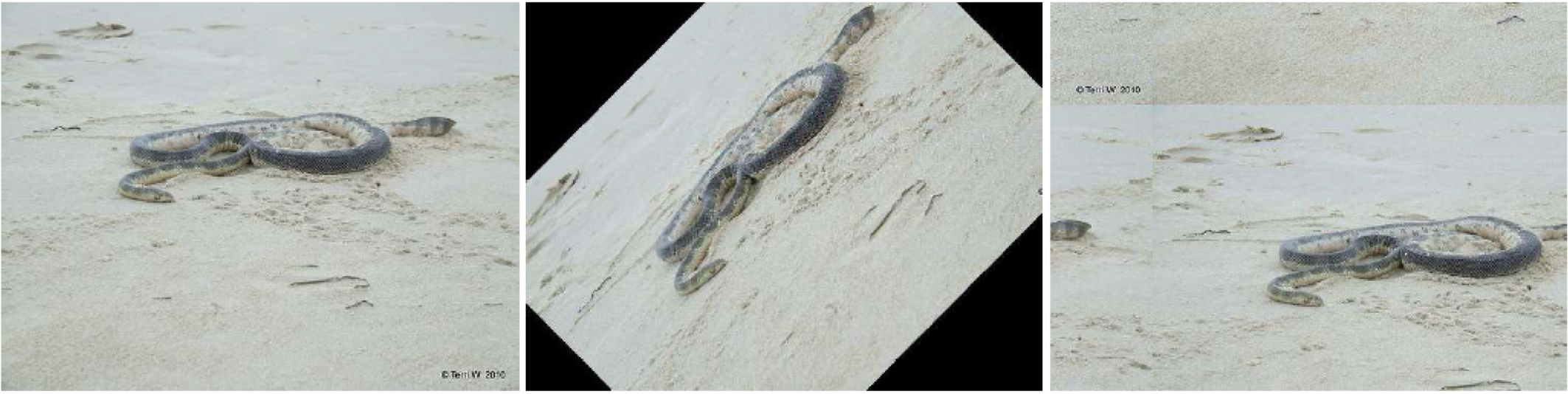}
        \caption{Input images.}
    \end{subfigure}
    \begin{subfigure}[t]{\linewidth}
        \centering
        \includegraphics[width=0.9\linewidth]{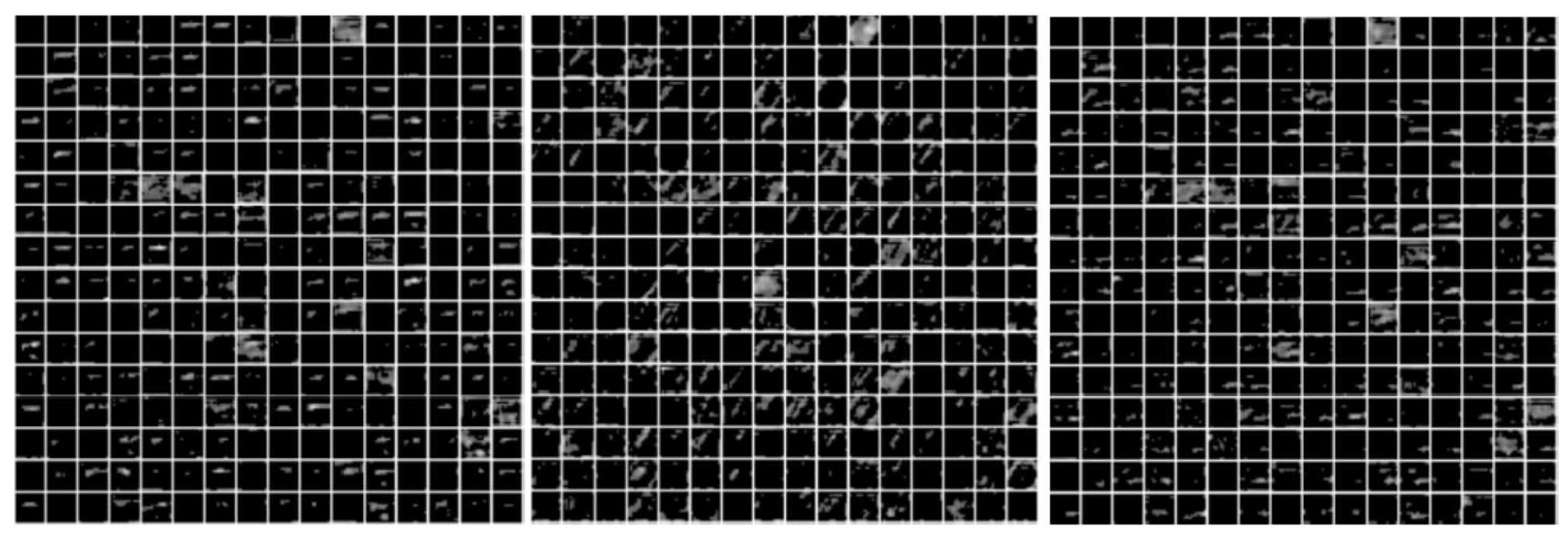}
        \caption{Output feature maps of Conv5 in AlexNet before patch reordering.}
    \end{subfigure}
    \begin{subfigure}[t]{\linewidth}
        \centering
        \includegraphics[width=0.9\linewidth]{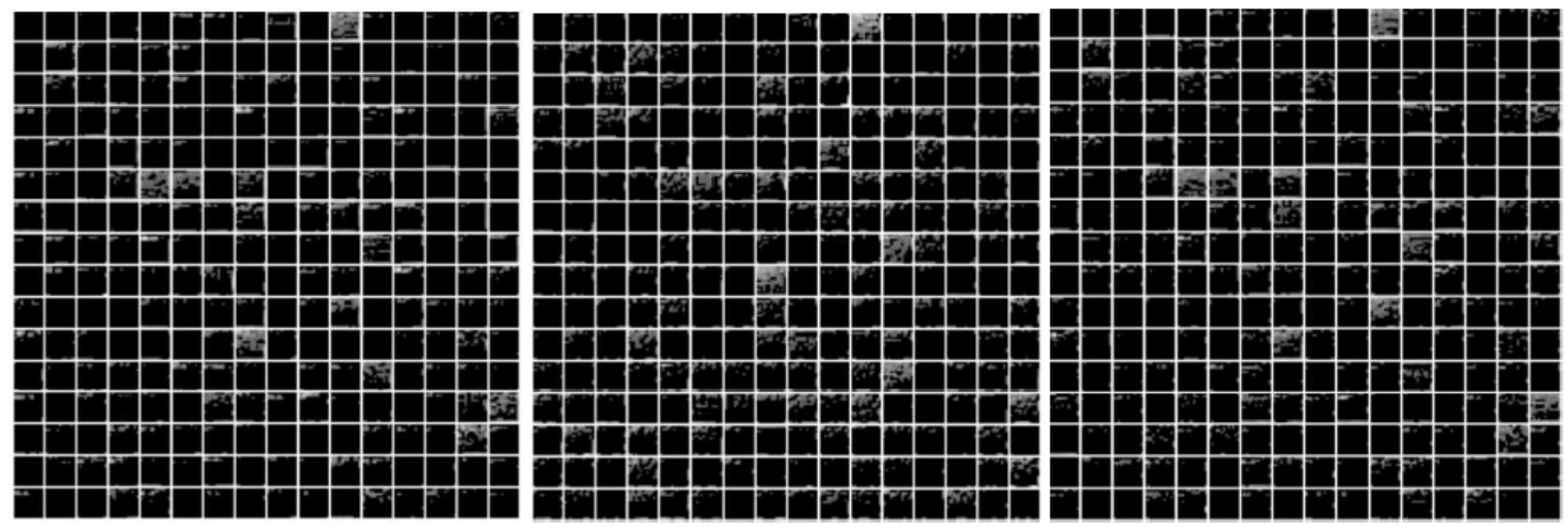}
        \caption{Output feature maps of Conv5 in AlexNet after patch reordering.}
    \end{subfigure}
    \caption{The feature maps of conv5 for the first image in the ImageNet 2012
        validation set. We can see that with patch reordering, the feature map
        is much more consistent than original CNN when faced with global
        rotations and translations.}
    \label{fig:representations}
\end{figure*}

To investigate the effect of applying patch reordering on the representations
of transformed images, we show the output feature maps of Conv5 in Alexnet in
Fig. \ref{fig:representations}. We can see that with patch reordering, the feature map
is much more consistent than original CNN when faced with global
rotations and translations.

\subsection{Effect of Patch Reordering on Different Layers}

To investigate the effect of applying patch reordering to different convolutional layers and the effect of pyramid levels, we train different PR-CNN models with patch reordering applied to $1\sim5$ convolutional layers with level $1$ or $2$. When $l=1$, we divide the feature maps into $4\times 4$ blocks.
For $l=2$, the feature maps are first divided into $2\times 2$ blocks, and each block is further divided into $2\times 2$ sub-blocks. The experimental results are presented in Fig. \ref{fig:effect}. We can see that the performance drops significantly when we perform patch reordering in low layers. Meanwhile, multi-level reordering does not result in a significant difference to single- level reordering in regard to higher convolutional layers. Low-level features, such as edges and corners, are detected in low layers, and they must be combined in a local spatial range to conduct further recognition. Because patch reordering breaks this local spatial correlation and treats each block as an independent feature, the generated representation becomes less meaningful. This explanation can also clarify the phenomenon that the multi-level division of feature maps significantly improves model performance in lower layers because a hierarchical reordering will preserve more local spatial relationships than will a single one.

\section{Conclusion}
In this paper, we introduce a very simple and effective way to improve the rotation and translation invariance of CNN models. By reordering the feature maps of CNN layers, the model is relieved from encoding location invariance into its parameters. Meanwhile, CNN models are able to generate more consistent representations when faced with location changes of local patterns in input. Our architecture does not need any extra parameters or pre-processing on input images.
Experiments show that our model outperforms CNN models in both image recognition and image retrieval tasks.

\textbf{Acknowledgments} This work is supported by NSFC under the contracts No.61572451 and No.61390514, the 973 project under the contract No.2015CB351803, the Youth Innovation Promotion Association CAS CX2100060016, Fok Ying Tung Education Foundation WF2100060004, the Fundamental Research Funds for the Central Universities WK2100060011, Australian Research Council Projects: FT-130101457, DP-140102164, and LE140100061.

\bibliographystyle{aaai}
\bibliography{ref}

\end{document}